\pdfoutput=1

\documentclass[11pt]{article}

\usepackage[]{ACL2023}

\usepackage{times}
\usepackage{latexsym}

\usepackage[T1]{fontenc}

\usepackage[utf8]{inputenc}
\usepackage{booktabs}
\usepackage{microtype}
\usepackage{multirow}
\usepackage{inconsolata}
\usepackage{subcaption}
\usepackage{xspace}

\newcommand{\OURS}{\textsc{LEGENT}\xspace}

\newcommand{\legentlogo}[1][2.9em]{%
  \raisebox{-0.4\height}{\includegraphics[height=#1]{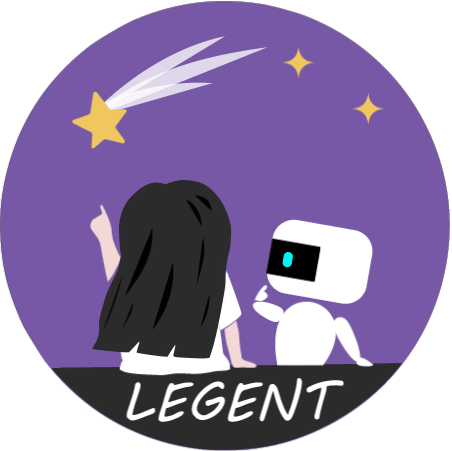}}
}

\usepackage{amssymb}

\usepackage{listings}
\PassOptionsToPackage{table}{xcolor}
\definecolor{codegreen}{rgb}{0,0.6,0}
\definecolor{codegray}{rgb}{0.5,0.5,0.5}
\definecolor{codepurple}{rgb}{0.58,0,0.82}
\definecolor{backcolour}{rgb}{0.99,0.99,0.98}
\lstdefinestyle{mystyle}{
    backgroundcolor=\color{backcolour},
    commentstyle=\color{codegreen},
    keywordstyle=\color{magenta},
    numberstyle=\tiny\color{codegray},
    stringstyle=\color{codepurple},
    basicstyle=\ttfamily\footnotesize,
    breakatwhitespace=false,
    breaklines=true,
    captionpos=b,
    keepspaces=true,
    numbers=left,
    numbersep=5pt,
    showspaces=false,
    showstringspaces=false,
    showtabs=false,
    tabsize=2,
    frame=shadowbox,
    rulesepcolor=\color{red!20!green!20!blue!20},
    xleftmargin=1em,xrightmargin=0em,aboveskip=1em,
    framexleftmargin=1em,
}

\usepackage{makecell}
\usepackage{graphicx}
\usepackage{enumitem}
\usepackage{booktabs} 
\usepackage{graphicx} 

\title{\legentlogo\hspace{3pt}\OURS: Open Platform for Embodied Agents}

\author{
  Zhili Cheng, Jinyi Hu, Zhitong Wang, Shengding Hu, An Liu,\\
  \textbf{Yuge Tu, Pengkai Li, Lei Shi, Zhiyuan Liu, Maosong Sun\thanks{
    \ \ Corresponding author. Email: sms@tsinghua.edu.cn}
  } \\
  Department of Computer Science and Technology, Tsinghua University
  \\
  \texttt{\{chengzl22, hu-jy21, wangzt23,  hsd23\}@mails.tsinghua.edu.cn}\\
  \url{https://legent.ai}
  \vspace{5pt}
  }

\makeatletter
\let\@oldmaketitle\@maketitle%
\renewcommand{\@maketitle}{\@oldmaketitle%
  \vspace{-20pt}
  \centering
  \includegraphics[width=0.98\linewidth]{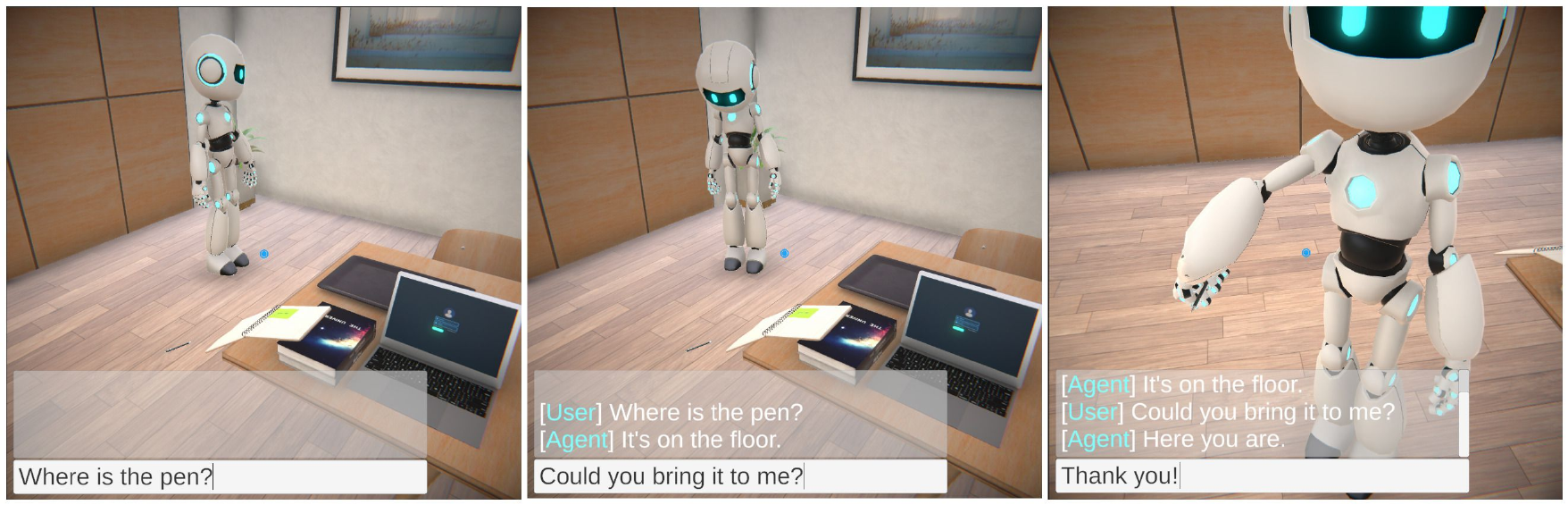}
  \vspace{-8pt}
  \captionof{figure}{
    Interaction with the embodied agent in \OURS. These sequential interactions showcase the agent's ability to answer the user's questions and follow the user's instructions.
  }
  \label{fig:crown}
  \vspace{17pt}
 }
\makeatother

\begin{document}

\maketitle

\begin{abstract}
Despite advancements in Large Language Models (LLMs) and Large Multimodal Models (LMMs), their integration into language-grounded, human-like embodied agents remains incomplete, hindering complex real-life task performance in physical environments. Existing integrations often feature limited open sourcing, challenging collective progress in this field. We introduce \OURS, an open, scalable platform for developing embodied agents using LLMs and LMMs. \OURS offers a dual approach: a rich, interactive 3D environment with communicable and actionable agents, paired with a user-friendly interface, and a sophisticated data generation pipeline utilizing advanced algorithms to exploit supervision from simulated worlds at scale. In our experiments, an embryonic vision-language-action model trained on \OURS-generated data surpasses GPT-4V in embodied tasks, showcasing promising generalization capabilities.
\end{abstract}

\section{Introduction}
\label{sec:intro}
Large Language Models (LLMs) \cite{brown2020language,achiam2023gpt, touvron2023llama, touvron2023llama2} and Large Multimodal Models (LMMs) \cite{openai2023gpt4v, team2023gemini, liu2024visual, hu2024large} present inspiring capabilities in understanding and generating human-like text and realistic images. However, their direct application in embodied AI, where agents interact in physical or simulated environments, is still primitive. LLMs and LMMs lack the necessary grounding \cite{harnad1990symbol} in physical interactions to operate in these settings effectively.

Research in embodied intelligence has evolved significantly, leading to more realistic and sophisticated environments \cite{kolve2017ai2,puig2018virtualhome,savva2019habitat, puig2023habitat} and increasingly challenging tasks \cite{das2018embodied,gordon2018iqa,batra2020rearrangement,yenamandra2023homerobot}.
However, these traditional environments and approaches are typically incompatible with current LLMs and LMMs, which hinders the seamless integration of task execution via language interaction. Consequently, these approaches do not leverage the extensive generalizable knowledge present in LLMs and LMMs.

To achieve generalizable embodied intelligence, two key factors are crucial: language grounding to utilize the extensive knowledge in LMMs, and the expansion of training data for embodied AI. 
There have been noteworthy efforts in combining embodied AI with LMMs~\cite{reed2022generalist, brohan2023rt}. They collect large-scale training data from embodied scenes and train end-to-end models that interpret both language and visual inputs and perform corresponding actions. However, the lack of open-source access to these environments and datasets restricts open-source community-wide progress in this field. Therefore, the academic community urgently requires an open platform that facilitates the integration of language grounding in embodied environments and schemes to generate large-scale training data for embodied agents based on LLMs and LMMs.

Towards this aspiration, we introduce \OURS, an open and user-friendly platform that enables scalable training of embodied agents based on LLMs and LMMs. \OURS contains two parts. First, it provides a 3D embodied environment with the following features: (1) Diverse, realistic, and interactive scenes; (2) Human-like agents with egocentric vision capable of executing actions and engaging in direct language interaction with users; (3) User-friendly interface offering comprehensive support for researchers unfamiliar with 3D environments. 
Second, \OURS builds a systematic data generation pipeline for both scene generation and agent behavior, incorporating state-of-the-art algorithms for scene creation~\cite{deitke2022️, yang2023holodeck} and trajectory generation. In this way, extensive and diverse trajectories of agent behavior with egocentric visual observations and corresponding actions can be generated at scale for embodied agent training.

To demonstrate the potential of \OURS, we train a basic vision-language-action model based on LMMs with generated data on two tasks: navigation and embodied question answering. The model processes textual and egocentric visual input and produces controls and textual responses directly. The prototype model outperforms GPT-4V~\cite{openai2023gpt4v}, which lacks training in an embodied setting. The generalization experiment reveals the \OURS-trained model's ability to generalize to unseen settings. \OURS platform and its documentation are publicly available at {\url{https://legent.ai}}.

\section{Related Work}

\textbf{Embodied Environment.} Embodied environments are extensively utilized in games \cite{johnson2016malmo, oh2016control, beattie2016deepmind} and robotics \cite{kolve2017ai2, yan2018chalet, xia2018gibson, gan2020threedworld, li2021igibson, puig2023habitat3}, with a primary focus on visual AI and reinforcement learning. Some platform focuses on specific embodied tasks, such as manipulation \cite{yu2020meta,makoviychuk2021isaac}, navigation \cite{chang2017matterport3d,dosovitskiy2017carla}, or planning-oriented agents \cite{puig2018virtualhome, shridhar2020alfworld, wang2022scienceworld}. 
However, the environment setups and data frameworks of existing platforms fall short in accommodating the training of LMMs. LMMs excel in the supervised learning paradigm and necessitate diverse and large-scale data to integrate embodied capability. 
Existing platforms are not yet ready to scale, including: the primarily supported reinforcement learning methods require careful reward engineering, the diversity of the training data cannot be easily expanded, and collecting data for imitation learning on these platforms requires manual effort.

\textbf{LMMs-based Embodied Agent.} Noteworthy
studies have concentrated on developing embodied
models capable of end-to-end operation, as demonstrated in the works of ~\citet{reed2022generalist,brohan2023rt,belkhale2024rt}. However, the datasets and models in these
studies are not publicly available.

\textbf{Scene Generation.} Scene generation has demonstrated significant effectiveness in training embodied agents by ProcTHOR~\cite{deitke2022️}. Compared to employing manually crafted rules used in ProcTHOR, recent studies \cite{wen2023anyhome, yang2023holodeck, feng2024layoutgpt} leverage prior knowledge of LLMs and propose algorithms to generate diverse, high-quality scenes.

\textbf{Agent Trajectory Generation.} Some research focuses on crafting reward functions to guide small policy models \cite{yu2023language,xian2023towards,wang2023robogen,ma2023eureka}.
However, there will be huge costs and instability when applying reward-based training to large foundation models. 
Meanwhile, pioneering efforts have been made in code generation for robotics \cite{liang2023code,singh2023progprompt, vemprala2023chatgpt, huang2023voxposer} and trajectory generation for imitation learning \cite{garrett2021integrated, kamath2023new, dalal2023imitating}. These efforts align with our approach to generating large-scale embodied trajectories for training LMMs.


\section{\OURS}
In this section, we introduce our platform \OURS. The design of \OURS involves scene, agent, and interface. All three components are specially tailored for the integration of LLMs and LMMs, and ensure scalability.

\begin{figure}[t]
    \includegraphics[width=0.95\linewidth]{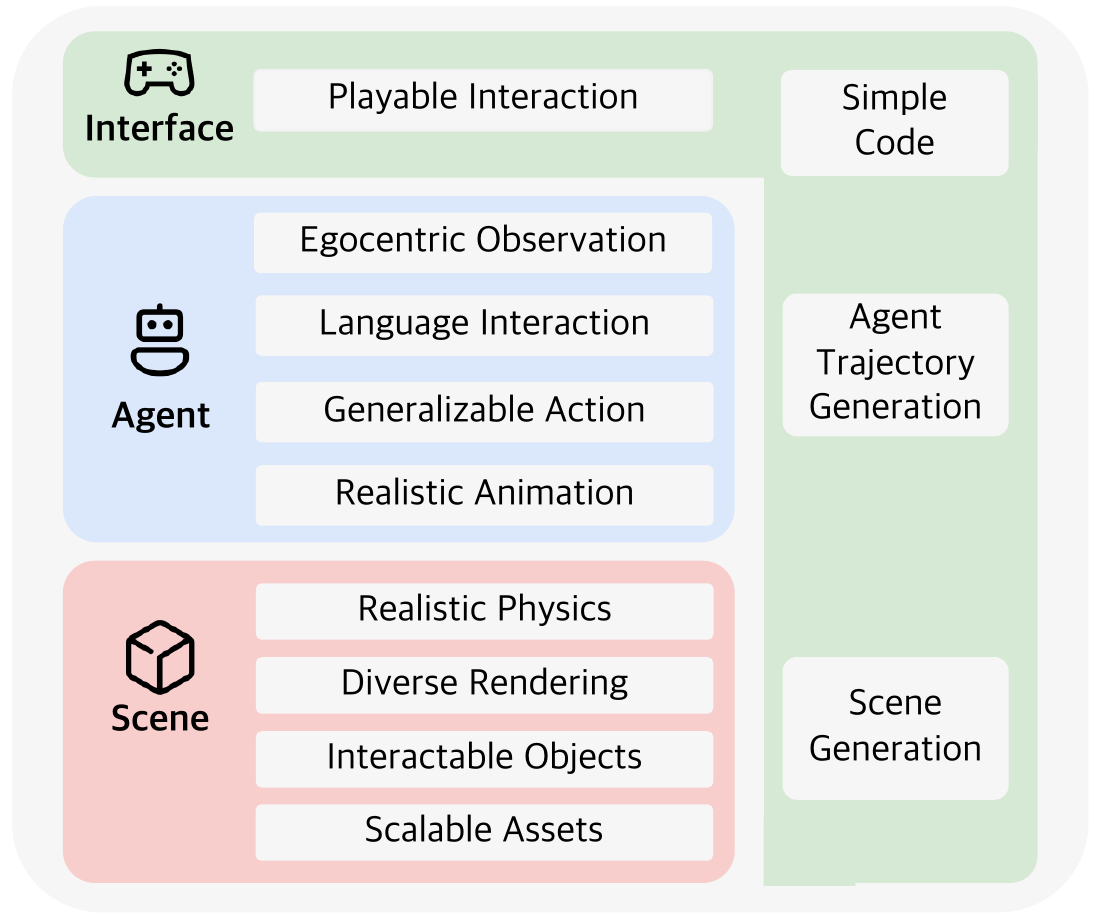}
    \caption{Features of \OURS.}
    \label{fig:mainfigure}
    \vspace{-1em}
\end{figure}

\subsection{Scene}

The design of the scenes in \OURS emphasizes \textbf{interactivity} and \textbf{diversity}, striving for a versatile and scalable environment that enriches the training of embodied agents for wide application.

\textbf{Realistic Physics.}  \OURS provides a real-time simulation that closely mirrors real-world physics based on game engines. It supports realistic effects like gravity, friction, and collision dynamics, improving agents' embodied comprehension or aiding the development of generative world simulators~\cite{yang2023learning}.

\textbf{Diverse Rendering.} \OURS introduces another facet of generalization via diverse rendering. Unlike the fixed stylized renderings in games and the emphasis on photorealism in robotics, \OURS integrates these styles by customizing the rendering functions, which allows easy transitions between rendering styles to accommodate different requirements for flexible usage.

\textbf{Interactable Objects.} In \OURS, both agents and users can manipulate various fully interactable 3D objects, which enables actions such as picking up, transporting, positioning, and handing over these objects. Additionally, the environment supports interaction with dynamic structures, such as doors and drawers. We anticipate that the scope of these dynamic structures will be significantly broadened through the application of generative methods~\cite{chen2023urdformer}.

\begin{figure}[t]
    \includegraphics[width=0.95\linewidth]{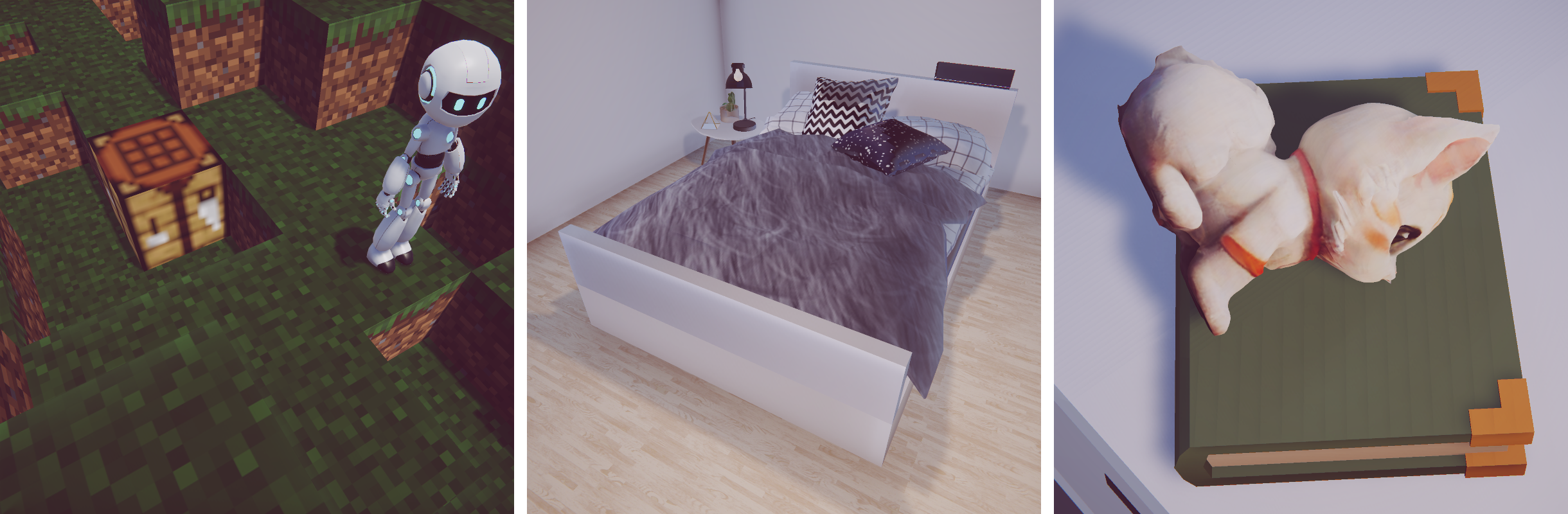}
    \caption{Examples of importing external assets: user-supplied assets (left); existing datasets (middle); assets generated by generative models (right).}
    \label{fig:asset}
    \vspace{0em}
\end{figure}

\textbf{Scalable Assets.} \OURS supports importing customized objects at runtime, including user-supplied 3D objects, objects from existing datasets~\cite{deitke2023objaverse} and those created by generative models~\cite{siddiqui2023meshgpt,wang2024crm}, as illustrated in Fig.~\ref{fig:asset}. We choose glTF as the import format for its openness and broad compatibility. This feature grants users the flexibility to customize the scene by strategically placing these assets or integrating them seamlessly into scene generation algorithms.

\subsection{Agent}
\label{agent} 
The agent is designed with two criteria: emulating human interactions and compatibility with LMMs. 

\textbf{Egocentric Observations.}
Following the previous study for interactive embodied agents \cite{team2021creating}, the agent is equipped with egocentric vision. The egocentric vision is captured by mounting a camera on the agent's head. 

\textbf{Language Interaction.}
Users and agents can communicate with each other in natural language in \OURS. Grounding language within the environment has the potential to connect the extensive knowledge in LLMs and LMMs with embodied experience.

\textbf{Generalizable Actions.}
Agents in \OURS are capable of performing a range of actions, including navigation, object manipulation, and communication.
Regarding the instantiation of actions, existing literature can be broadly categorized into two types: \textit{executable plans}~\cite{puig2018virtualhome,shridhar2020alfworld} and \textit{control}~\cite{kolve2017ai2,savva2019habitat}. In \textit{executable plans}, actions are expressed through sub-steps to complete a task, such as ``\textit{walk towards apple 1}'', which depends on internal states and annotations for execution, or requires an additional neural executor module compatible with a planning module~\cite{driess2023palm}. \textit{Control}, on the other hand, refers to the action expression like ``\textit{move forward 1 meter, rotate to the right 30 degrees}'', which is considered more generalizable. In \OURS, we use \textit{control}, targeting generalizing to new environments including real-world settings. The learned actions can be integrated with diverse actuators with the least additional effort.

\begin{table}[t]
\scalebox{0.85}{
\begin{tabular}{l|ll}
\toprule
 Actions       & Description \\ \midrule
Speak & Send a message. \\
Move* & Move forward by a specified distance. \\
Rotate* & Adjust the view horizontally or vertically. \\
Interact & Grab, put, open, or close targeted objects.              \\ \bottomrule
\end{tabular}}
\caption{List of actions in \OURS. * means the action is continuous (meters or degrees).}
\label{tab:actionlist}
\end{table}

Another important action design is allowing the agent to execute continuous actions such as moving forward across a continuous distance, as opposed to moving in a grid-by-grid manner.
This design offers two advantages for LMMs: (1) It minimizes the inference cost of LMMs by eliminating the need for constant frame-by-frame inference. (2) It addresses the issue of minimal information gain observed when an agent moves incrementally in a stepwise manner, a process that creates less effective data for training large models. This design draws parallels to the use of keyframes in video processing and making direct training of autoregressive LMMs ~\cite{alayrac2022flamingo, awadalla2023openflamingo, lin2024vila} feasible.
Specifically, the actions currently supported in \OURS are shown in Table~\ref{tab:actionlist}.
Considering the current capability of LMMs, \OURS temporarily omits the complex control of agents' body joints. Adding these degrees of freedom to allow more flexible action will be explored in the future.

\begin{figure}[t]
\centering
    \includegraphics[width=0.98\linewidth]{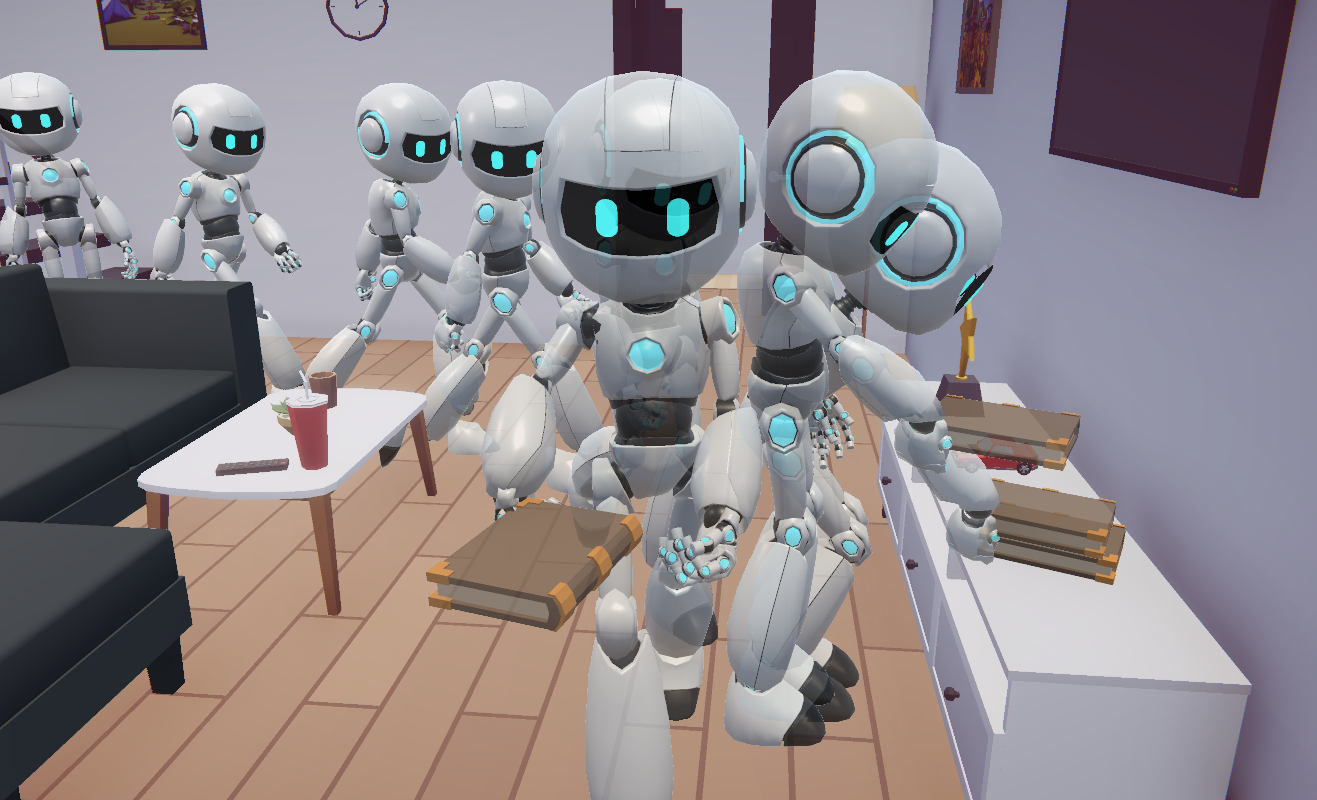}
    \caption{An example of humanoid animations, demonstrating accurate object grasping and body movement through spatial planning and inverse kinematics.}
    \label{fig:animation}
    \vspace{-1em}
\end{figure}

\textbf{Realistic Animation.} \OURS features precise humanoid animations using inverse kinematics and spatial algorithms, enabling lifelike movements, as shown in Fig.~\ref{fig:animation}. It is important for enhancing nonverbal interactions in AI systems and contributes to robotic control and text-to-motion research. Also, when combined with egocentric vision, it offers a cost-effective alternative for immersive experiences similar to Ego4D~\cite{grauman2022ego4d}, which requires a huge cost to collect.

\subsection{Interface}

Our platform offers a user-friendly interface for researchers to integrate LLMs and LMMs with the embodied environment easily, with little need for expertise in 3D environments. Detailed guidance is available in our documentation.

\textbf{Playable Interaction.} The user interface of \OURS is designed to be as intuitive as playing a video game with the agent within the environment, utilizing just a keyboard and mouse for navigation and interaction. This interface facilitates straightforward visual debugging and qualitative analysis and simplifies the process of conducting hands-on demonstrations. 

\textbf{Simple Code.}
\OURS is equipped with a Python toolkit to enable the interaction between the agent and the environment.
 The coding interface of our Python toolkit is simple, with concise code examples available in our documentation.

\textbf{Scene Generation Interface.} Our platform incorporates various scene-generation techniques. Currently, we support methods including procedural generation and LLM-based generation. 
We provide a straightforward JSON format for specifying a scene, enabling users to easily develop their own scene generation methods.

\textbf{Agent Trajectory Generation Interface.} We offer an agent trajectory generation interface specifically designed for training LMMs. Using this interface, users can create training datasets that consist of egocentric visual records and corresponding ground truth actions paired with task instructions or queries, as elaborated in Section~\ref{traj}.

\textbf{Hardware Requirements.}  \OURS is cross-platform. It can run effortlessly on personal computers without demanding particular prerequisites or complex setups, and it facilitates connections to remote servers for training and deployment, thus enhancing its accessibility.

\section{Data Generation}

The second part of \OURS is a scalable data generation pipeline. It aims at exhaustively exploiting the inherent supervision from simulated worlds and supporting large-scale training of general-purpose embodied agents. 
Here we elaborate on the implementation of our data generation framework.

\subsection{Scene Generation}

Scene generation offers agents with diverse embodied experiences. \OURS has currently integrated two scene generation methods: (1) Procedure generation efficiently creates large-scale scenes. (2) Language-guided generation captures the semantics of textual queries and leverages common sense knowledge to optimize spatial layouts.

\textbf{Procedural Generation.} We utilize the procedural generation algorithm created by ProcTHOR~\cite{deitke2022️}, designed to create realistic indoor scenes at scale by integrating prior knowledge of object placement and spatial relationships. 
The implementation process starts with drafting a house layout, followed by the placement of large furniture, and ends with the arrangement of small objects. During the process, spatial algorithms are used to prevent object overlap and ensure precise placement.
We provide an interface that allows users to input specific conditions for object occurrence and placement, enabling the generation of scenes tailored to specific tasks. In addition, instead of employing human annotators as previous work does, we utilize LLMs for asset annotation, establishing an efficient \textit{automatic asset annotation} pipeline that facilitates future asset expansion.

\textbf{Language Guided Generation.} We implement methods in Holodeck~\cite{yang2023holodeck} into \OURS and offer an LLM-powered interface to generate single or multi-room indoor scenes given any natural language query.
This process resembles procedural generation but is driven by LLMs instead of human-written programs.
Instead of using the depth-first-search solver in Holodeck, we ask LLMs to determine the exact locations of doors and floor objects, granting LLMs more control over the room layout. Collision detection is used to prevent interference between objects during generation.

\begin{figure}[t]
\centering

\begin{subfigure}[b]{1\linewidth}
  \centering
  \includegraphics[width=0.44\linewidth]{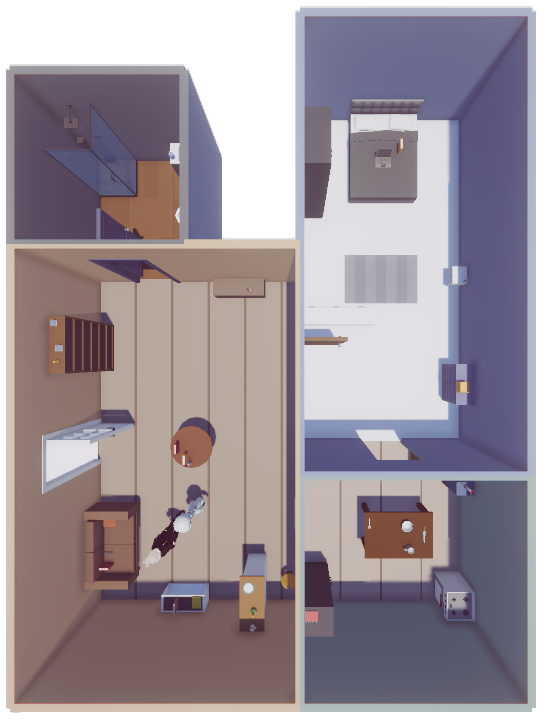}
  \hspace{-0.01\linewidth}
  \includegraphics[width=0.54\linewidth]{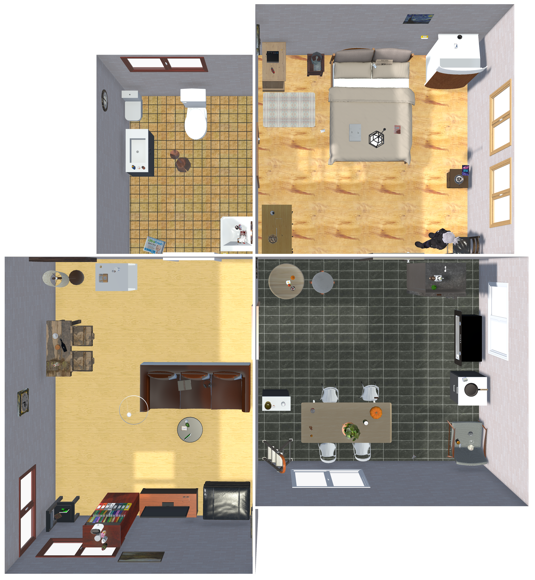}
\end{subfigure}

\caption{Examples of generated scenes.}
\label{fig:groups}
\vspace{-1em}
\end{figure}

\subsection{Task Generation}

\label{task}
We create diverse tasks expressed in language paired with specific scenes, thereby contextualizing each task within the environment. We employ the following two strategies for task generation.

\textbf{Task Generation for Given Scenes.}  In this strategy, we serialize the generated scenes into a detailed textual description and present it to LLMs with crafted instructions. LLMs assume the role of human users, generating a variety of tasks. This approach is especially effective for generating diverse tasks automatically.

\textbf{Scene Generation for Given Tasks.} This approach efficiently generates large-scale samples for specific tasks based on the scene generation algorithm. For instance, when the task involves querying an object's location, the algorithm generates a scene that includes the object and its receptacle, inherently creating question-answering annotations. As shown in Table \ref{tab:tasktemplates}, we provide some basic task templates that are ideal for creating large-scale scenes, which are particularly useful for pretraining \textit{fundamental capabilities} of embodied control, spatial comprehension, and basic language grounding across diverse scenes.

\begin{table}[t]
\scalebox{0.85}{
\begin{tabular}{l|l}
\toprule

 Task       & Intermediate Code \\ \midrule
 come here  & goto\_user()         \\ 
 go to \textit{A}    & goto(\textit{a})  \\
 pick up \textit{A}  & goto(\textit{a}) target(\textit{a}) interact()        \\ 
 bring me \textit{A} & goto(\textit{a}) target(\textit{a}) interact() goto\_user()                \\ 
 where is \textit{A} & find(\textit{a}) speak(\textit{C})                \\ 
 put \textit{A} on \textit{B} & goto(\textit{a}) target(\textit{a}) interact()   \\
 & goto(\textit{b}) target(\textit{b}) interact() \\
 \bottomrule
\end{tabular}}
\caption{Currenly provided task templates and intermediate code templates. 
\textit{A} is the object's name, and \textit{a} is the object's environment identifier. \textit{C} denotes the name of the receptacle on which \textit{a} is placed.}
\label{tab:tasktemplates}
\vspace{-1em}
\end{table}

\subsection{Trajectory Generation}
\label{traj}

Trajectories for training embodied agents comprise continuous sequences of egocentric observations and actions. The main challenge lies in accurately determining ground-truth actions for each step.

We use LLMs and \textit{motion planners} to label the ground truth actions. Inspired by pioneering works in code generation for robotics, we utilize LLMs to write intermediate codes from provided state descriptions and instructions. These codes are instantiated as motion planners, designed to calculate the optimal actions at each step given the 
internal states of the environment. Each motion planner operates in a step-by-step manner in the environment, with visual observations collected during the process. This approach is consistent with the concept of Task and Motion Planning (TAMP)~\cite{garrett2021integrated} in robotics, where the LLMs and the motion planners respectively fulfill the functions of task planning and motion planning.

\begin{figure}[t]    
    \begin{center}{
    \includegraphics[width=0.96\linewidth]{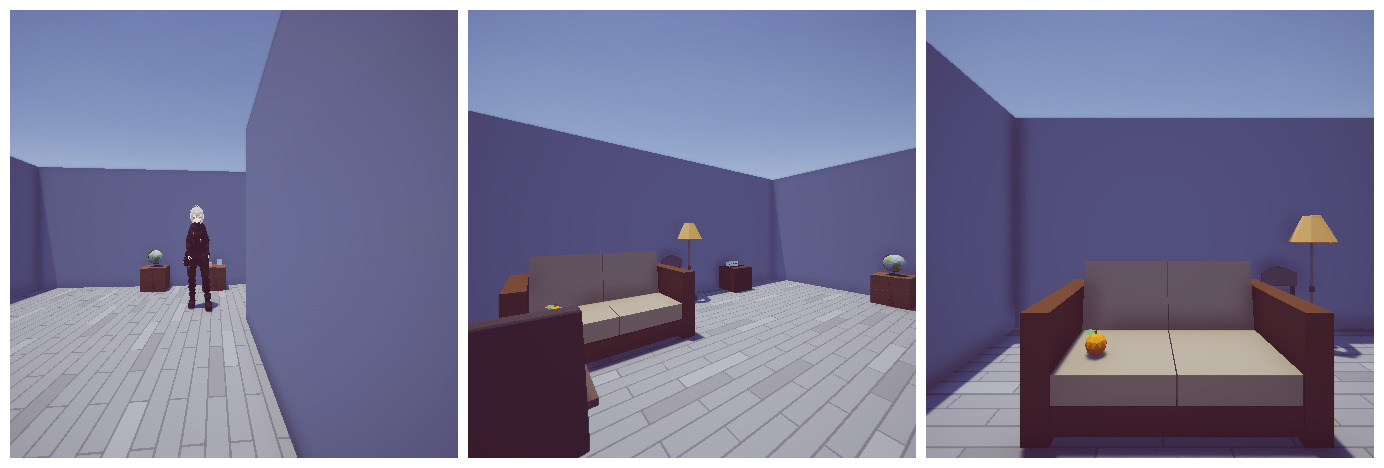}
    \caption{A generated trajectory for task ``Where is the orange''. The actions for the three observations are: 
    1. \textit{rotate\_right(-59)};
    2. \textit{move\_forward(1.2), rotate\_right(-35)};
    3. \textit{speak("It's on the sofa.")}.
    \label{fig:arch}}}
    \end{center}
    \vspace{-1em}
\end{figure}

We demonstrate this process using an example task ``\textit{Where is the orange?}''. As shown in Figure \ref{fig:arch}, to finish the task, the agent needs to search the room and answer the question.
LLMs map the task to the appropriate code usage, determine the object identifier of the orange  in the scene, and recognize its placement from the state description, thereby generating the following intermediate code:

\begin{lstlisting}[language=Python, label={code:kopl-2}]
find(36) # object identifier of orange
speak("It's on the sofa.")
\end{lstlisting}

Note that the code-writing is annotation-oriented.
Even though LLMs can directly answer the question from the state description, it still invokes ``\textit{find}''. Then the code ``\textit{find}'' is instantiated as a motion planner that utilizes pathfinding algorithms~\cite{hart1968formal} incorporating visibility checks. The pathfinding algorithm calculates the waypoints of the shortest path from the agent to the target object using a navigation mesh. The motion planner then calculates the controls of the agent to navigate along these waypoints. For instance, in the first observation shown in Figure \ref{fig:arch}, the agent needs to rotate 59 degrees to the left to orient to the next waypoint, resulting in the action ``\textit{rotate\_right(-59)}''. Similarly, in the second observation, the agent needs to perform certain actions to move to the subsequent waypoint. This motion planner concludes when the target object enters the agent's field of view. \OURS records visual observations and actions during this process as a trajectory, which can be exported as a video or an image-text interleaved sequence. The actions use a unified code representation, compatible with the outputs of LMMs.

Similar to ``\textit{find}'', each intermediate code is designed with the ability to generate optimal controls using the internal world states. In addition, each task template mentioned in Section~\ref{task} is equipped with intermediate code templates, as shown in Table~\ref{tab:tasktemplates}, eliminating the need for LLMs in large-scale data generation for specific tasks.

\subsection{Prototype Experiments}

\begin{figure}[t]    
    \begin{center}{
    \includegraphics[width=1.0\linewidth]{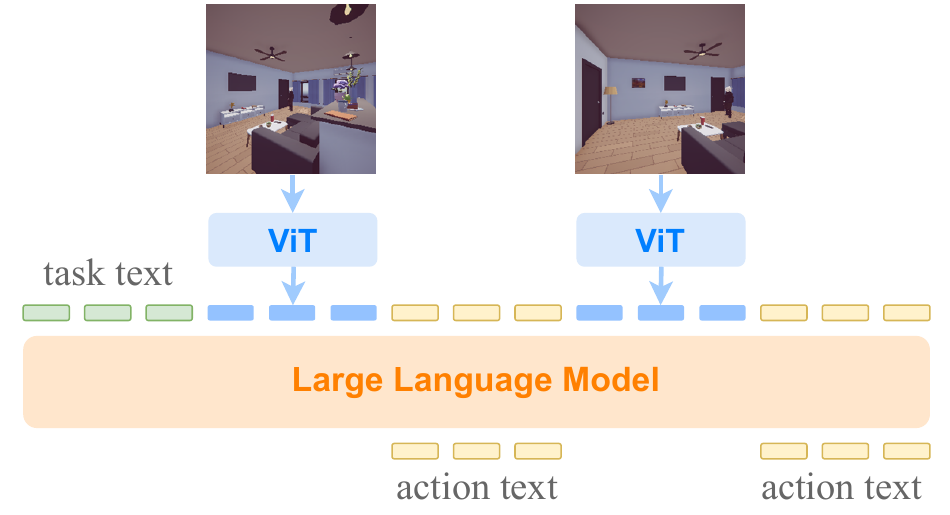}
    \caption{The vision-language-action model architecture used in the prototype experiments.}
    \label{fig:model}}
    \end{center}
\end{figure}

We conduct a prototype experiment to assess the utility of generated data on two embodied tasks: ``Come Here'' for navigation and ``Where Is'' for embodied question answering \cite{das2018embodied}. Task complexity varied from navigating in one room to the more intricate two rooms. We generate 1k and 10k trajectories for the initial three tasks (``Come Here'' in one or two rooms and ``Where Is'' in one room) and assess the models on 100 trajectories across all four tasks. The ``Where Is'' task in the two-room setting serves as a generalization test, which is not included in the training data. 

Due to the lack of powerful video understanding models, we temporarily only use the observation at the end of each continuous action, formulating one trajectory as an image-text interleaved sequence. We utilize VILA-7B \cite{lin2024vila} as our backbone due to its capability in interleaved inputs. As illustrated in Fig.~\ref{fig:model}, we train the vision-language-action (VLA) model to predict current action based on task descriptions and interleaved context of previous observations and actions,.

The results presented in Table \ref{tab:exp} lead to several observations: (i) GPT-4V struggles in these tasks, reflecting a lack of embodied experience in mainstream LMMs. (ii) Increasing training data improves the model performance. (iii) The navigational skills developed from the ``Come Here'' task in a two-room environment generalize well to the untrained task scenario, enhancing the model's ability to navigate in two rooms for the embodied question answering task. We leave the exploration of more large-scale training in the future work.

\begin{table}[t]
\centering
\scalebox{0.90}{
\begin{tabular}{l|cc|cc}
\toprule
Task              & \multicolumn{2}{c|}{Come Here}   & \multicolumn{2}{c}{Where Is}   \\ \midrule
Room Num          & \multicolumn{1}{c|}{One}  & Two  & \multicolumn{1}{c|}{One}  & Two*  \\ \midrule
GPT-4V (zero-shot) & \multicolumn{1}{c|}{0.21} & 0.17 & \multicolumn{1}{c|}{0.25} & 0.22 \\ \midrule
VILA-7B-Sep 1K    & \multicolumn{1}{c|}{0.87} & 0.28 & \multicolumn{1}{c|}{0.30}  & 0.22 \\ 
VILA-7B-Sep 10K   & \multicolumn{1}{c|}{\textbf{0.96}} & \textbf{0.70} & \multicolumn{1}{c|}{\textbf{0.94}} & 0.52 \\ 
VILA-7B-Joint     & \multicolumn{1}{c|}{\textbf{0.96}} & \textbf{0.70} & \multicolumn{1}{c|}{0.92} & \textbf{0.65} \\ \bottomrule
\end{tabular}}
\caption{Success rates on two embodied tasks. \textit{VILA-Sep} denotes models fine-tuned separately for each task, whereas \textit{VILA-Joint} refers to models trained jointly on both tasks. * means generalization test. }
\label{tab:exp}
\vspace{-1em}
\end{table}

\subsection{Demo of \OURS}
The demo video of \OURS is available at the link\footnote{\url{https://video.legent.ai}}, which is partially shown in Fig.~\ref{fig:crown}.
The demonstration exemplifies the engagement with embodied agents in \OURS, primarily leveraging LLMs and motion planners described in Section~\ref{traj}. With advancements in LMMs' capability of egocentric perception and control, we foresee the evolution of this demonstration into a fully embodied experience, independent of any extra internal information. We will also pursue this goal by further scaling the data generation for model training.

\section{Conclusion and Future Work}

In this work, we present \OURS, an open platform for developing embodied agents, focusing on integrating LMMs with scalable embodied training. By bridging the gap between embodied AI and LMM's development, we hope \OURS inspires research in this field. We are committed to the ongoing development of \OURS, making it more scalable and user-friendly. In our future releases, we prioritize: (1) Building a more diverse data generation pipeline. (2) Scaling model training. (3) Unifying humanoid animation with robotic control and refining the physics to make actions more applicable to the real world. (4) Improving scene generation and integrating text-to-3D and image-to-3D methods to support more diverse and realistic scenes.





\bibliography{anthology,custom}
\bibliographystyle{acl_natbib}

\end{document}